\documentclass[10pt,twocolumn,letterpaper]{article}

\usepackage{cvpr}
\usepackage{times}
\usepackage{epsfig}
\usepackage{graphicx}
\usepackage{amsmath}
\usepackage{amssymb}

\usepackage{subfig}
\usepackage{array} 
\usepackage{graphicx,multirow}
\usepackage{caption}

\usepackage{booktabs}

\usepackage[dvipsnames]{xcolor}
\DeclareRobustCommand{\legendsquare}[1]{%
  \textcolor{#1}{\rule{1ex}{1ex}}%
}

\usepackage[pagebackref=true,breaklinks=true,letterpaper=true,colorlinks,bookmarks=false]{hyperref}

\cvprfinalcopy 

\def\cvprPaperID{3818} 

\ifcvprfinal\fi
\begin{document}

\title{







Exploiting Class Similarity for Machine Learning with Confidence Labels and Projective Loss Functions

}

\author{Gautam Rajendrakumar Gare\\
Carnegie Mellon University\\
Pittsburgh, USA\\
{\tt\small gautam.r.gare@gmail.com}
\and
John Michael Galeotti\\
Carnegie Mellon University\\
Pittsburgh, USA\\
{\tt\small jgaleotti@cmu.edu}
}


\maketitle

\begin{abstract}
  

Class labels used for machine learning are relatable to each other, with certain class labels being more similar to each other than others (\eg images of cats and dogs are more similar to each other than those of cats and cars). Such similarity among classes is often the cause of poor model performance due to the models confusing between them. Current labeling techniques fail to explicitly capture such similarity information. In this paper, we instead exploit the similarity between classes by capturing the similarity information with our novel confidence labels. Confidence labels are probabilistic labels denoting the likelihood of similarity, or confusability, between the classes. Often even after models are trained to differentiate between classes in the feature space, the similar classes' latent space still remains clustered. We view this type of clustering as valuable information and exploit it with our novel projective loss functions. Our projective loss functions are designed to work with confidence labels with an ability to relax the loss penalty for errors that confuse similar classes. We use our approach to train neural networks with noisy labels, as we believe noisy labels are partly a result of confusability arising from class similarity. We show improved performance compared to the use of standard loss functions. We conduct a detailed analysis using the CIFAR-10 dataset and show our proposed methods' applicability to larger datasets, such as ImageNet and Food-101N.

\end{abstract}

\section{Introduction}

\newlength{\Awidth}
\setlength{\Awidth}{0.75 in}
\newlength{\Aheight}
\setlength{\Aheight}{0.68 in}

\begin{figure}[!ht] 
\centering

\setlength{\tabcolsep}{1pt} 
\def\arraystretch{0.1} 

\newcolumntype{C}{>{\centering\arraybackslash}m{\Awidth}<{}}
\newcolumntype{F}{>{\centering\arraybackslash}m{0.1\Awidth}<{}}
\resizebox{\columnwidth}{!}{
\begin{tabular}{F C C}

&
\subfloat{\tiny Cross-entropy Loss} &
\subfloat{\tiny Log-Projection Loss} \\[-1.5ex]

\rotatebox[origin=c]{90}{\centering \tiny 0\% noise ratio } &
\subfloat{\includegraphics[height = \Aheight, width = \Awidth]{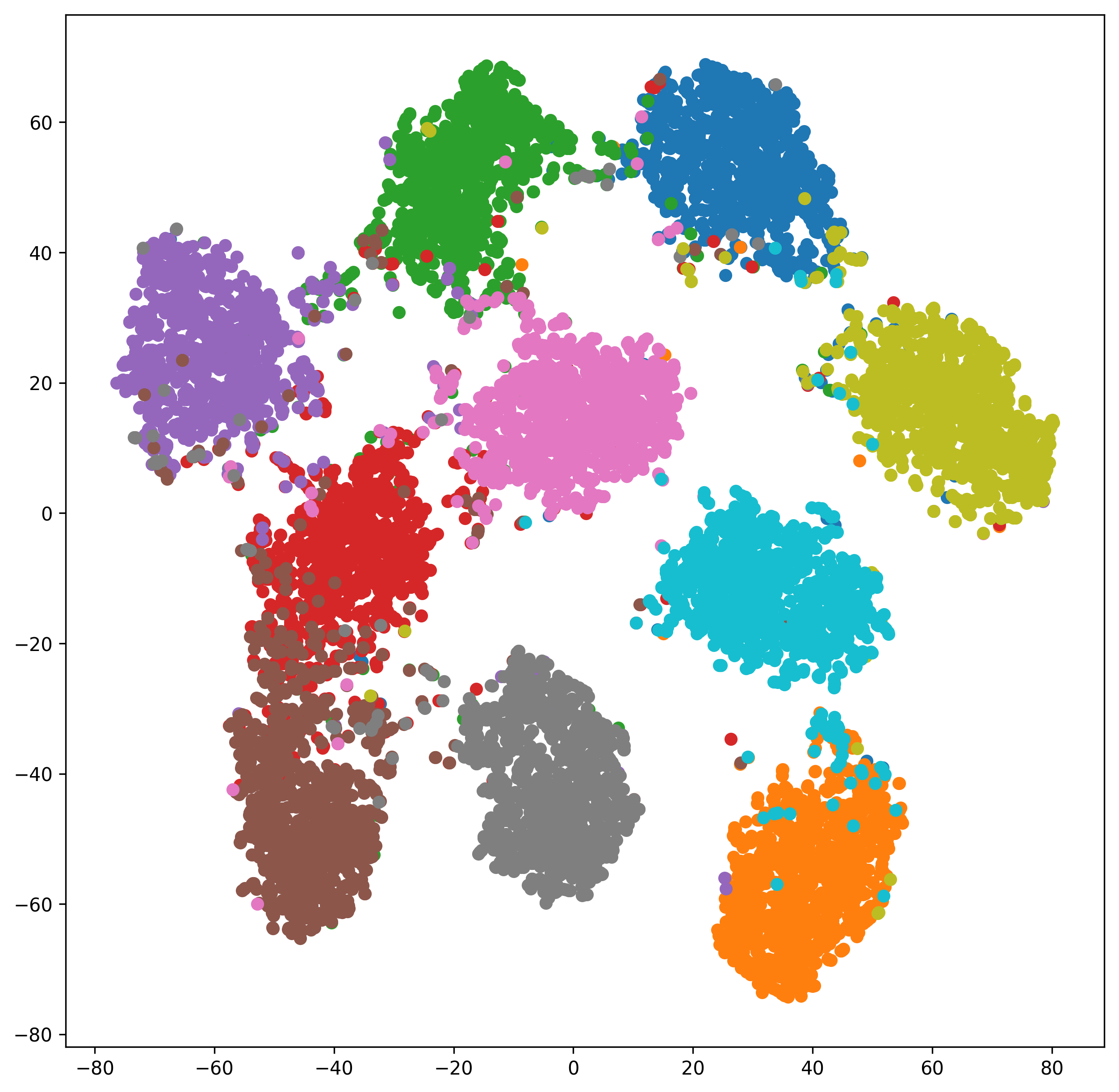}} &
\subfloat{\includegraphics[height = \Aheight, width = \Awidth]{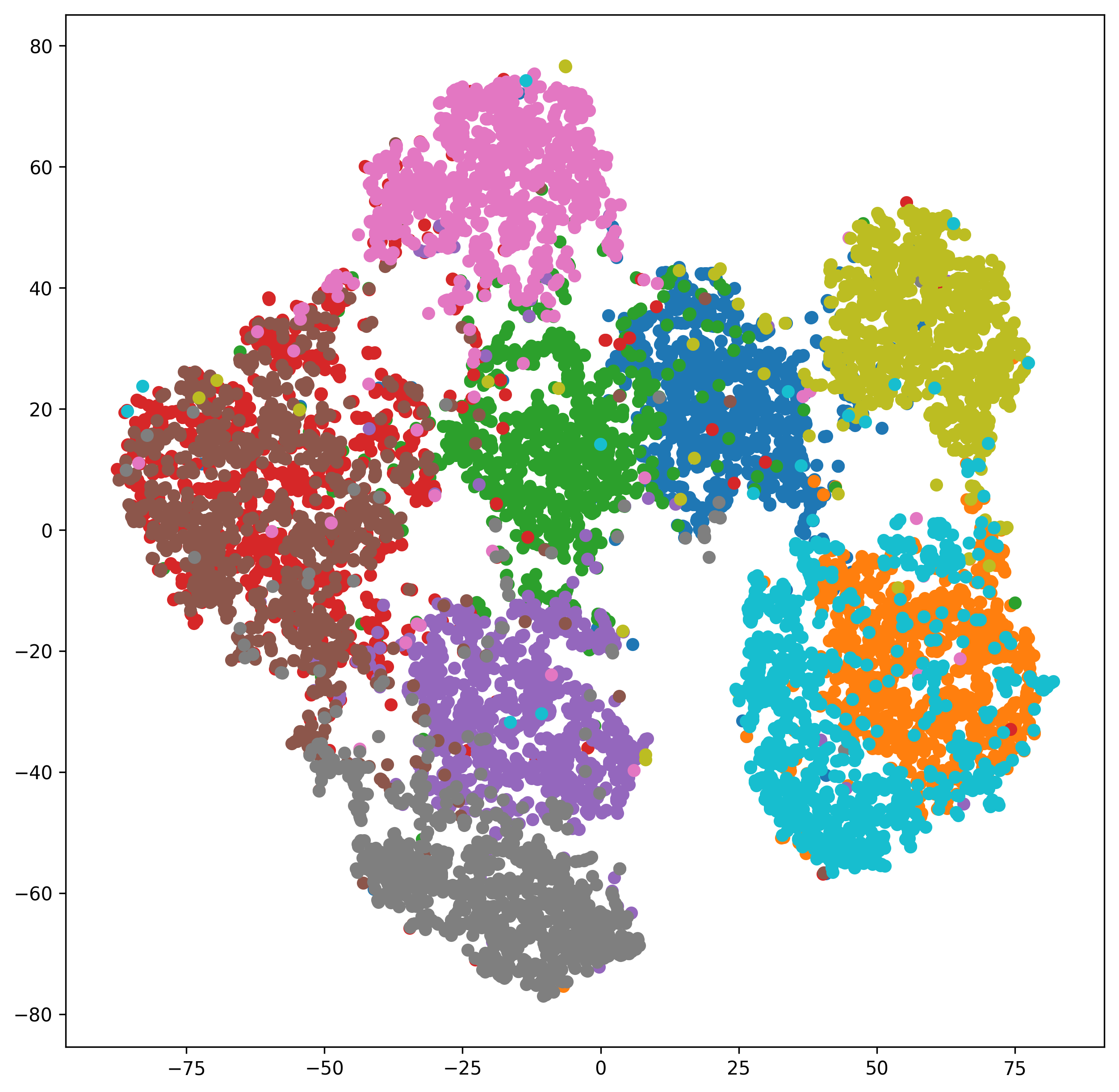}} \\[-1.5ex]

\rotatebox[origin=c]{90}{\centering \tiny 80\% noise ratio } &
\subfloat{\includegraphics[height = \Aheight, width = \Awidth]{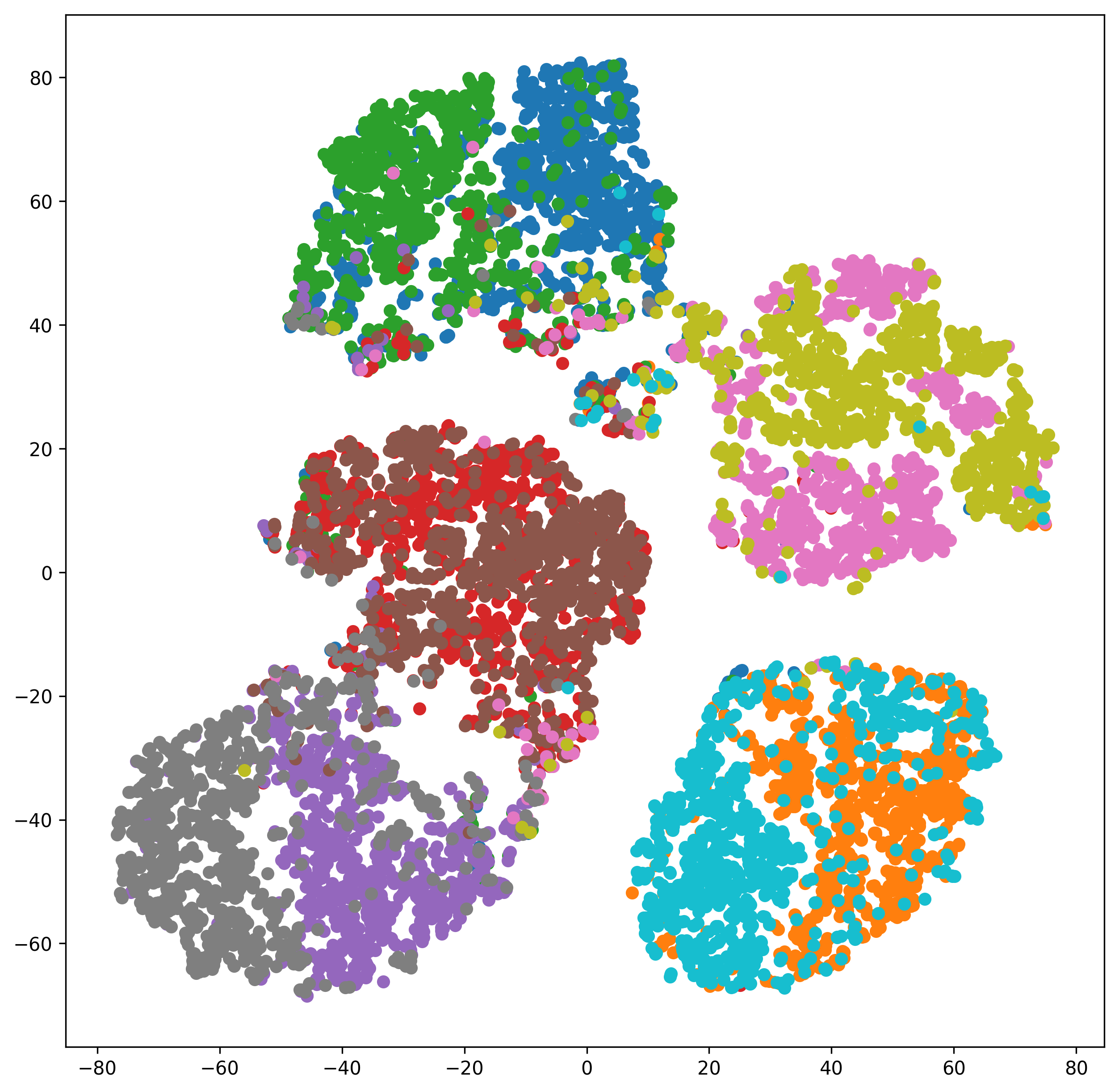}} &
\subfloat{\includegraphics[height = \Aheight, width = \Awidth]{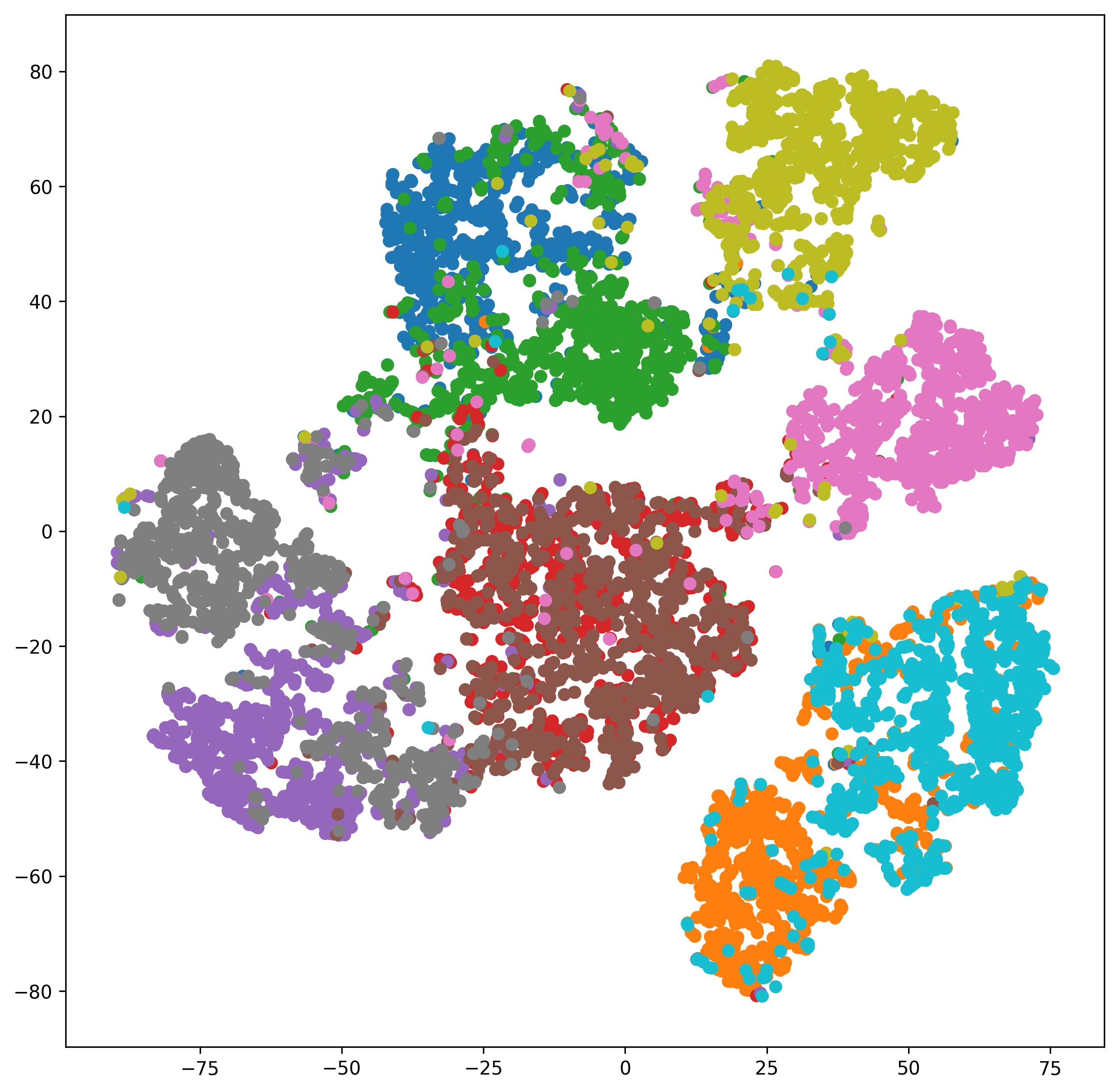}} \\[-1.5ex]

\multicolumn{3}{c}{\subfloat{\includegraphics[width = 0.5\columnwidth]{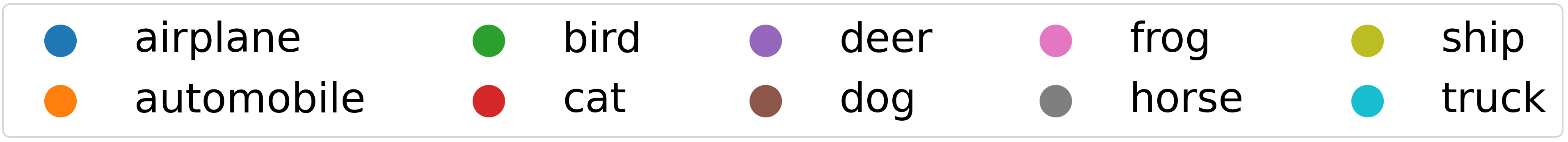}}} \\ 

\end{tabular}

}
\caption{
\small The t-SNE plots of trained models on CIFAR-10 using Cross-entropy and Log-Projection loss, with 0\% and 80\% asymmetric noise ratio and trusted set $M$ = 1k. Although cross-entropy attempts to learn feature embeddings that differentiate classes, we observe that similar classes still end up relatively closer. This suggests that \emph{AI models have a natural tendency to learn features that cluster according to class similarities}. A model trained instead with \textbf{our proposed Log-Projection loss exploits rather than fight this natural clustering}, irrespective of the noise.
}
\label{fig:class_distribution}
\end{figure}


\newlength{\width}
\setlength{\width}{0.75 in}
\newlength{\height}
\setlength{\height}{0.68 in}

\begin{figure}[!ht] 
\centering

\setlength{\tabcolsep}{1pt} 
\def\arraystretch{0.1} 

\newcolumntype{C}{>{\centering\arraybackslash}m{\width}<{}}
\newcolumntype{F}{>{\centering\arraybackslash}m{0.1\width}<{}}
\resizebox{\columnwidth}{!}{
\begin{tabular}{F C F C}

&
\subfloat{\tiny Cross-entropy Loss} &
&
\subfloat{\tiny Log-Projection Loss} \\[-1.5ex]

\rotatebox[origin=c]{90}{\centering \tiny Accuracy (\%) } &
\subfloat{\includegraphics[height = \height, width = \width]{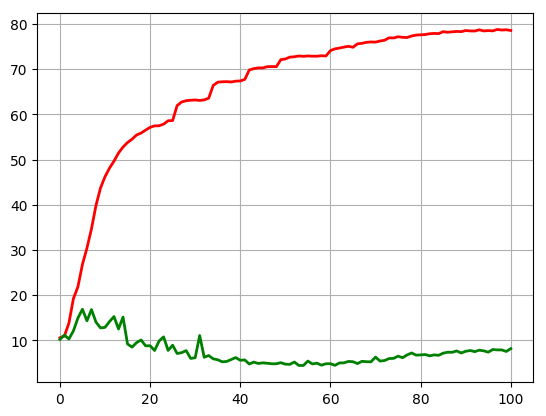}} &
\rotatebox[origin=c]{90}{\centering \tiny Accuracy (\%) } &
\subfloat{\includegraphics[height = \height, width = \width]{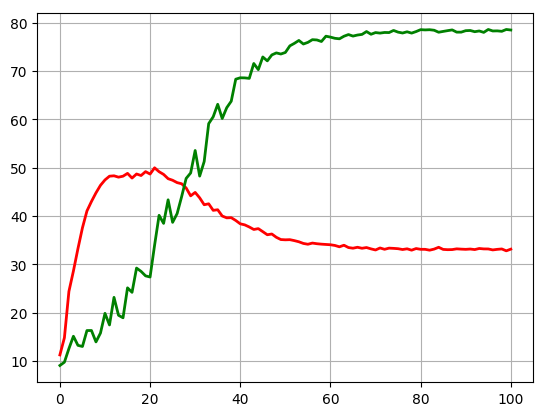}} \\[-1.5ex]

&
\subfloat{\tiny Epochs} &
&
\subfloat{\tiny Epochs} \\[-0ex]

\end{tabular}

}
\caption{
\small Image classification results on CIFAR-10 with 80\% asymmetric noise ratio and trusted set $M$ = 1k. Labels:
\legendsquare{Red} noisy train 
\legendsquare{Green} evaluation.
The model trained using confidence labels and Log-Projection loss auto-corrects, whereas the cross-entropy loss leads to model overfitting on the noisy training data.
}
\label{fig:training_curves}
\end{figure}

Objects are relatable to each other. A cat bears more resemblance to a dog than to a car. Such distinctions that come easily to us humans help us understand even unseen objects, but such similarities can also be the cause of confusion. Can you distinguish rabbits from hares? Current AI systems are not immune to confusability arising from class similarities. \cite{imagnet_class_similarity_main} and \cite{imagnet_class_similarity} have shown that the most confused classes on the ImageNet dataset follow the WordNet hierarchy, implying that class similarity is a contributing factor to poor model performance.

Machine learning systems are usually not provided any supervision on the inter-class similarity, as the typical labels used for machine learning do not explicitly capture such class similarities. This leaves the models to implicitly discern the class similarity information from the training data to distinguish between classes, leading to poor performance. Several works \cite{soft_labels, focal_loss}, have looked at this problem by treating it as a regularization issue. \cite{soft_labels} presented soft labels, a label regularization strategy wherein a fixed value is assigned to all the non-target classes, but using soft labels has been shown to actually lead to loss of class similarity information \cite{label_smoothing_hurts}. 

We introduce confidence labels that explicitly model the inter-class similarity relations and thus provide models the supervision to learn such relations. Confidence labels are probabilistic labels denoting the likelihood of similarity (or confusability) among the classes. We define them as a vector of (real or pseudo) probabilities of each possible label, which is analogous to a neural-network classifier's final softmax activations.  In the present work, we obtain confidence labels on a per-class basis, either though heuristic measures or by using pretrained models. (Future work will explore applying unique confidence labels per image or even per pixel.) Confidence labels are an easy 
way of introducing a-priori inter-class similarity information into the neural network training, which, when coupled with our novel projective loss functions, encourages both preserving and learning the naturally occurring class distributions.

Most of the typical machine learning objective functions try to learn features that semi-equally separate all the classes into different clusters in the latent feature space, ideally pulling apart all the clusters to be orthogonal to each other. Implicit in prior approaches is the assumption that classes are not only distinguishable, but that with the right features, each class is equally distinguishable from every other class. However, these systems still end up having tight clusters for similar classes, which is suggestive of a naturally-occurring distribution arising from inter-class similarity \cite{label_smoothing_hurts}, leading us to suspect that any attempt at ``overcoming'' these naturally occurring distributions may be futile. We further suspect that modifying the label space to be orthogonal to each other may be incorrect, especially when working on open-set classes \cite{openset_learning}. A model that is trained on a closed set of classes might not appropriately fit an unseen class into its label space when operating under the assumption that every class is distinct.



To preserve the naturally occurring class similarity distribution, we introduce projective loss functions tailored to work with confidence labels with an ability to relax the loss penalty for similar classes. Our projective loss functions preserve and reinforce the naturally occurring cluster distributions, such that similar classes stay closer while still separating dissimilar classes. In contrast, typical loss functions (e.g. cross-entropy) impartially force all class clusters to move apart (pushing toward orthogonality), without regard to inter-class similarity. Figure-\ref{fig:class_distribution}, we observe that similar classes (\eg cat and dog) end up being relatively closer irrespective of the label distribution trained upon. 

Similarly, contrastive learning approaches \cite{supervised_contrastive_learning}
handle inter-class similarity by trying to bring the feature embeddings of same-class instances closer while moving away from feature embeddings of different classes.  This approach is predicated on the assumption of inter-class dissimilarity, thus ignoring the existence of inter-class similarity. We believe our approach is the first to explicitly exploit the inter-class similarity.


We test our approach to modeling the label space in the challenging setting of noisy labels (mislabelled data). We show that the mere usage of our projective loss combined with our confidence labels, without yet incorporating many of the other standard training-enhancement techniques, can achieve comparable performance to the state-of-the-art systems when dealing with high label noise stemming form inter class similarity (i.e. asymmetric or semantic noise \cite{asymmetric_label_noise}).

\textbf{Contributions:}
Our primary novel ideas are that (A) class-similarity information should be exploited during training, that (B) similar classes need not, and should not, be well separated  in latent feature space, and that (C) noisy labels often occur as a result of class similarity, which would lead naturally occurring label noise to be asymmetrically clustered.

Our  main  technical  contributions are (1) Introducing our novel Confidence Labels as a new way of labelling data based on probabilistic labels that denote the likelihood of similarity. (2) Methods for the creation and generation of confidence labels that instil class similarity prior to model training. (3) A novel family of Projective loss functions tailored to handle confidence labels, which reinforce naturally occurring class similarity distributions/relations. (4) Training mechanisms for working with confidence labels and projective loss functions.

Our simple, but effective, strategy explicitly exploits class similarity information for training more robustly with noisy labels.  It achieves significant robustness at high asymmetric noise levels on CIFAR-10 \cite{cifar_dataset} comparable to current state-of-the-art methods, while being a straightforward and simple strategy and not increasing training time as compared to current methods. We also show improvements on large-scale datasets, ImageNet \cite{ImageNet} and Food101-N \cite{food101n_dataset_lee2017cleannet}, without yet incorporating many of the ``standard'' training-enhancement techniques such as AutoAugment \cite{autoaugment} and Cosine learning rate decay with warm restarts \cite{cosine_learning_rate}.

\section{Related Work}
\label{}

\textbf{Soft labels}, which are k sparse labels (k $\neq$ 1) have been used in several contexts as an alternate to one-hot labels. \cite{soft_labels} presented soft labels as a label regularization strategy wherein a single fixed value is assigned to all the non-target classes. This can be considered as a special case of our proposed confidence labels, where the inter class similarity is unknown. \cite{label_smoothing_hurts} showed that using such regularized soft labels leads to the loss of class similarity information.  (As we show in figure \ref{fig:class_distribution}, our method preserves class similarity information). \cite{network_prediction_as_soft_labels} made use of the network predictions as soft labels for their domain and task transfer learning, in order to better align source domain labels with target domain labels. \cite{human_scenicness_as_soft_labels} trained models to match human rated ``scenicness'' scores, viewed as soft labels. \cite{age_as_soft_labels} used soft labels to encode the randomness of ageing and proposed a ``soft softmax'' loss function to work with these soft labels. As compared to these prior works, our confidence labels can be viewed as soft labels encoding the label class similarity information, where in we present ways to exploit this similarity information for label-noise-robust learning. 




\textbf{Noisy labels.} Working with noisy labels has been extensively researched \cite{teacher_student_learning_method, google_label_weighting_relabelling, label_noise_robustness, food101_dataset, food101n_dataset_lee2017cleannet, food101n_setup_han2019deep}, with the most common techniques attempting to determine which training samples are wrongly labeled in order to regularize or downweight the loss of the mislabelled samples \cite{loss_correction1, loss_correction_approach_patrini2017making} or to determine the correct label by using a trusted set of correctly labeled data \cite{google_label_weighting_relabelling, meta_learned_soft_labels_vyas2020learning, food101n_setup_han2019deep}. 
Several mentor-student curriculum approaches have also been proposed \cite{teacher_student_learning_method, Li2020ProductIR} where the mentor sets a curriculum for the student to learn for working with noisy labels. CleanNet \cite{food101n_dataset_lee2017cleannet} uses transfer learning to address noisy labels by learning features on a trusted set. \cite{loss_correction1} proposed a meta-learning algorithm that uses a trusted set to reweight noisy samples


The label correction approaches have been show to work well even at high noise levels \cite{google_label_weighting_relabelling, meta_learned_soft_labels_vyas2020learning}. \cite{food101n_setup_han2019deep} proposed a self-learning approach to work with noise labels, wherein a model is trained to classify and perform label correction in an iterative manner. \cite{google_label_weighting_relabelling} presented an effective technique of Distilling Effective Supervision(DES) using a small high-quality ``probe set'' to weight and relabel noisy data for supervised training, achieving significant improvements in accuracy when working with high noise levels. Unfortunately, their technique requires 4x more training time compared to normal training. \cite{meta_learned_soft_labels_vyas2020learning} presented a meta learning strategy to dynamically learn soft labels. It can be seen that the instance labels that emerge from their learning approach have the same structure as the confidence labels that we independently developed and propose. Their approach is one of learned label refinement, for which they continue to use standard cross-entropy loss when training and learning/refining their soft labels.
%
%


In certain contexts it may be dangerous to automatically refine or relabel the training data, especially using the primary-training error metric.
In particular, \emph{some safety critical applications such as medicine or autonomous vehicles may contain rare but important outliers that must be preserved, even if they statistically appear to have been mislabeled.}  We argue that preserving outlier labels (even if there is uncertainty as to whether they were correctly labeled) is important for long-tail real world distributions, where there will be correctly labelled outliers such as patients with rare diseases that confound their symptoms, downed power lines on the road ahead, or anything under the category of ``if it can happen, it will.''

\textbf{Robust loss functions.} 
Previously many loss functions have been designed that introduce some form of regularization for robust learning. \cite{focal_loss} presented Focal Loss, which downweights the loss of correctly classified samples such that the learning can take place for the misclassified samples. (Our proposed relaxation of the loss can be viewed in the same spirit.)
\cite{loss_correction_approach_patrini2017making} apply a loss correction for scenarios where the inter-class corruption probability is known. Similar to our approach, \cite{label_noise_by_loss_function} designed a loss function more capable of dealing with noisy labels. \cite{symmetric_cross_entropy_loss} designed a symmetric cross-entropy loss to address the shortcoming of cross-entropy loss wherein it overfits on easy classes and under-learns on hard classes, and they  applied their approach to work with noisy labels. Our proposed projective loss functions exploit the class similarity relation for robust learning.




\cite{supervised_contrastive_learning} designed a loss function which acts on the feature embeddings such that feature embedding of objects belonging to same class are closer than feature embeddings of different class objects. Our loss, takes this a step further where in addition to making same class embedding closer we make the feature embeddings of similar classes (not same, ex cat \& dog) closer than those of dissimilar classes (such as car).

In contrast to the prior work, our approach is simple, does not increase training time, and robustly learns from noisy labels automatically without resorting to label refinement or relabeling. Although our approach benefits from a trusted probe set, it still learns effectively when all training data is uncertain.  Our method does not require any additional/supplement learning, hyper-parameter tuning, etc.


\section{Proposed Method}
\label{}


\subsection{Confidence Labels}

We define confidence labels as probabilistic labels denoting the likelihood of similarity (or confusability) among the classes. It can be viewed as a probability-distribution vector of the class similarity scores of the object across labeling classes, with vector elements that sum to one. 

To introduce class similarity prior to model training, we propose to replace one-hot labels with these confidence labels. For simplicity when working with typical one-hot labeled images, we currently constrain all instances of a class to have the same confidence label, so that each one-hot vector has a corresponding confidence label.  Nothing prevents us from assigning independent, unique confidence labels to every image, more like the instance labels proposed in \cite{meta_learned_soft_labels_vyas2020learning}, but we argue that high-quality per-image confidence labels are better derived from the input of human experts. 


We propose the following simple methodology of defining per-class confidence labels based a one-hot labeled set.  Consider the case involving $C$ label classes, for two classes $a$ and $b$ under consideration $(a,b \in C)$, we assign a similarity score $S$ based on a heuristic H as:

\begin{equation}
    S_{ab} = H(a,b)
\end{equation}

Next we define the similarity group $G_a$ for class $A$, as the classes which have a similarity score $S$ greater than a threshold $\tau$. 

\begin{equation}
    G_a = [b \in C \text{ if } S_{ab} > \tau ]
\end{equation}

We define the confidence score $C_a(b)$ of $a$ for class $b$ as:


\begin{equation}
    \label{eq:C_a}
    C_a[b] = 
    \begin{cases}
        \text{softmax}(S_{ab}), & \text{if } b \in G_a\\
        0,              & \text{otherwise}
    \end{cases}
\end{equation}

Where the softmax is taken over the similarity group $G_a$. 

Confidence label $T$ of class $a$ is the collection of these confidence scores $C_a$.

The heuristic $H$ can be intuitively defined as done in section \ref{cifar_exps} or determined using pretrained models, as shown in section \ref{imagenet}. 

It needs to be noted that the spread of the confidence score is restricted to a small group of similar classes, as confusion-inducing class similarity general exists between a small subset of classes $k$ \cite{imagnet_class_similarity_main}. So confidence labels are generally $k$ sparse labels, but they can be though of as a generic label definition.  When $k=1$ they become normal one-hot labels, wherein the class under consideration is totally dissimilar from every other class. We term this as \emph{hard confidence labels}. On the other hand the soft labels of \cite{soft_labels} is the confidence label wherein the class under consideration is equally similar to all other classes.

\subsection{Projective Loss functions}

We present a family of projective loss functions tailored for confidence labels, that reinforce the naturally occurring class similarity with the ability to relax the loss penalty for similar classes.
The projective loss function is based on the dot product a.k.a. the inner product, such that it tries to maximize the model prediction $P(\phi)$ projection onto the confidence label $T$, where $\phi$ is the model parameters which we want to optimize. We introduce a relaxation function $r(t)$ which scales the target labels in order to relax the loss penalization for class labels which are less confident and thus give the model the added flexibility to make predictions that deviate from the labeled ``truth,'' which is useful when working with noisy labels.


\textbf{Projection (P) Loss:}
\begin{equation}
    \mathcal{L}_P = max(0, <T_r, T_r> - <T_r, P(\phi)>)
\end{equation}

\begin{equation}
    \tfrac{d \mathcal{L}_P}{d \phi} = - <T_r,  \tfrac{d P(\phi)}{d \phi}>
\end{equation}

\textbf{Log-Projection (LP) Loss:}
\begin{equation}
    \begin{aligned}
     \mathcal{L}_{LP} = max&( 0, \log (\tfrac{<T_r, T_r>}{<T_r, P(\phi)>}))
    \end{aligned}
\end{equation}
\begin{equation}
    \tfrac{d \mathcal{L}_{LP}}{d \phi}  = - \tfrac{1}{<T_r, P(\phi)>} <T_r,  \tfrac{d P(\phi)}{d \phi}>
\end{equation}

\begin{equation}
where, T_r = r(T)
\end{equation}






To ensure numerical stability a small constant ($1e-08$) is added to the loss. 



\subsubsection{Loss Visualization}
\begin{figure}[!ht] 
\centering
\includegraphics[height=0.5\columnwidth, 
keepaspectratio]{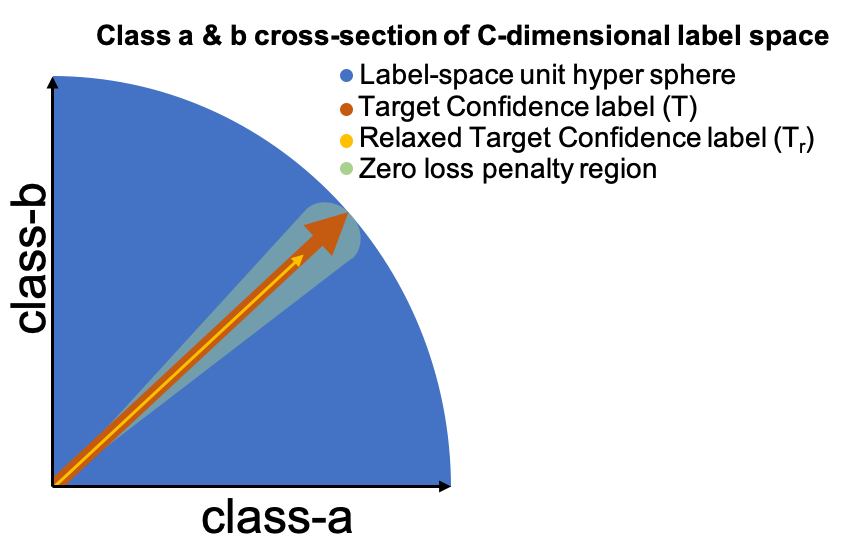} 
\caption{ \small Projection loss relaxation visualization. 
} 
\label{fig:projection_loss}
\end{figure}

Figure-\ref{fig:projection_loss} depicts the projection loss function's relaxation region around the relaxed target confidence label $T_r$. This gives the model the added flexibility to classify the object as either more like class-A or more like class-B, without incurring any loss penalty.

We use two relaxation functions, first the L2-norm function for training samples that might be inaccurately labeled, and second no-relaxation for the non-noisy trusted set. 

\begin{equation}
    r_{noisy}(t) = ||t||_2 t
\end{equation}

\begin{equation}
    \label{eq:no_relax}
    r_{trusted}(t) = t
\end{equation}


The relaxation lets the loss of closely aligned train samples go to zero, allowing only the gradients of the samples which are clearly misclassified to pass through. This can be related to the way Focal Loss  and Symmetric cross-entropy loss down-weight/regularize the gradients of the correctly classified samples \cite{focal_loss,symmetric_cross_entropy_loss}. 
The relaxation does raise a case where the loss goes to zeros and no further learning can happen.  This could be addressed by modifying the relaxation function $\tau$ (e.g. introducing a scaling parameter).  We do not evaluate this case here, leaving it for future exploration. 

Our projective loss function 
reinforces that classes of the same similarity group remain close in latent feature space while separating dissimilar classes. Unlike typical loss functions like cross-entropy loss which impartially forces all class clusters to move apart (ideally to be orthogonal), without regard to inter class similarity. Figure-\ref{fig:class_distribution}, 


Using the L2-norm, less-certain target confidence labels (with lower maximum confidence scores) are shrunk substantially and subsequently the model is meagerly penalized if swaying away from that training-data's uncertain target confidence label. Similarly, confidence labels with higher confidence scores are shrunk by a small factor, so  subsequently the model is heavily penalized if swaying away from that training-data's confident target label. 


The Projection loss can be related to Cosine-Embedding Loss, but unlike cosine embedding which tries to bring closer the embeddings of samples belonging to a class and pull apart the embedding of different classes, projection loss works on the label space, wherein it tries to closely align the model prediction class to more-confident (or typical one-hot encoded) confidence labels, with the added ability to relax the loss penalty for uncertain confidence labels.






\subsubsection{Repurposed Loss functions}
We can also repurpose existing loss functions to work with confidence labels. For instance, the cross-entropy loss (CE) can be repurposed to work with confidence labels by using the relaxed confidence label $T_r$ instead of normal one-hot labels, which we term \emph{projection cross-entropy (pCE) loss}. Similarly, substitution of $T_r$ for one-hot labels can allow other loss functions such as L1, MSE, Focal Loss, etc. to also use confidence labels.

\begin{equation}
    L_{pCE} = - \sum T_r log(P(\phi))
\end{equation}




\section{Experiments}
\label{exp}

We conduct a series of experiments on the CIFAR10 dataset.  We begin by testing various loss functions with confidence labels in section \ref{comparinng_loss_functions}, followed by experiments evaluating our approach on various noise settings in section \ref{noisy_label_exp}. We then go beyond CIFAR10, performing experiments on large-scale, real-world datasets in section \ref{real-world_exp}.

\subsection{CIFAR10 noisy label experiments}
\label{cifar_exps}

The CIFAR10 \cite{cifar_dataset} dataset consists of 10 class labels with certain class labels more similar than others. We manually defined similarity scores for the class labels following the same class-similarity groups $G_a$ as \cite{asymmetric_label_noise, class_dependent_noise}, which is TRUCK $\Leftrightarrow$ AUTOMOBILE, BIRD $\Leftrightarrow$ AIRPLANE, DEER $\Leftrightarrow$ HORSE, CAT $\Leftrightarrow$ DOG and FROG $\Leftrightarrow$ SHIP. We use the confidence score $C_a(a) = 0.6$ for the correct class $a$ and $C_a(b) = 0.4$ for the similar class $b$, for all CIFAR10 experiments in this paper unless indicated otherwise.


Similar to \cite{google_label_weighting_relabelling}, we make use of a small trusted set $M$ which is a subset of the training dataset $N$, where $M \subset N$ and $M << N$.  We are confident of the \emph{high quality labels} in $M$, so we assign higher confidence values to the trusted set, $C_a(a) = 0.95$ for the correct class $a$ and $C_a(b) = 0.05$ for the similar class $b$.  However, this subset is not used for any loss reweighting or label correction as conventionally used in similar prior setups \cite{google_label_weighting_relabelling}; rather, we simply include the more confident $M$ as part of the larger $N$ training mix. 

An initial (unreported) brief ablation study on labeling binary relations indicated that confidence scores of 0.6 vs 0.4 work well for noisy labels and that 0.95 vs 0.05 work well for high-quality labels. Accordingly, we used those values in the experiments presented here.


We follow the similar setup of symmetric and asymmetric noise for the CIFAR10 dataset as conventionally carried out in \cite{asymmetric_label_noise, loss_correction_approach_patrini2017making, google_label_weighting_relabelling}. Label transitions are parameterized by $r \in [0, 1]$ such that true class and wrong class have probability of $1 - r$ and $r$, respectively. For, the asymmetric noise the wrong class is limited to the similarity group $G_a$, whereas for the symmetric noise, all classes are likely. In the case of confidence labels, when the wrong class is selected, the confidence label of the wrong class is used in place of the correct-class confidence label.

We note that our experiment sets up confidence labels to be less informative than one hot labels for the correctness of the class. For a noise level $r$, the cross-entropy loss has $(1-r)N$ samples with high quality correct labels whereas our projective loss has only $M$ samples with high-quality correct labels and $(1-r)N - M$ low-quality correct labels.

\subsubsection{Implementation details and experimental setup}

For all the experiments on the CIFAR10 dataset, we evaluate using ResNet-34 implemented with PyTorch and trained using the stochastic gradient descent algorithm \cite{stochasticGradientUpdate} with a weight decay of $5e-4$, set with an initial learning rate of $0.2$. The model is trained on 4 Nvidia GeForce GTX 1080 Ti GPUs with a batch size of 256 for gradient update computation. The ReduceLRonPlateau learning-rate scheduler was used which reduces the learning rate by a factor (0.5) when the performance accuracy plateaus, with patience 5. The model is trained for 200 epochs. For the final evaluation, we pick the best model with the highest test-set accuracy. We report mean and standard deviation of the scores from \emph{3 independent runs} for all the experiments, similar to \cite{google_label_weighting_relabelling}. We follow \cite{supervised_contrastive_learning} for implementing ResNet \footnote{https://github.com/HobbitLong/SupContrast}.
%



To present a fair, base-line comparison, we do not use any advanced augmentation techniques such as AutoAugment \cite{autoaugment} apart from standard RandomResizedCrop (resized to 32x32) and RandomHorizonalFlip, nor do we use any specialized learning rate scheduler such as Cosine learning rate decay with warm restarts.  These and similar methods have been shown to boost the scores further in similar experimental setups \cite{google_label_weighting_relabelling}, and we suspect our methods would likewise benefit from future use of such advanced augmentation and learning-rate methodologies.

\subsubsection{Comparing loss functions for Confidence labels}
\label{comparinng_loss_functions}

We compare various standard loss functions under the asymmetric noise setting. We compare the performance of cross-entropy (CE) using one-hot labels with L1 loss, Mean-squared-error (MSE) loss, repurposed cross-entropy loss (pCE) and the proposed projection (P) loss and log-projection (LP) loss, all using the confidence labels instead of one-hot labels, with a trusted set $M$ of 1k, signifying 100 high quality confidence labels for each of the 10 class. 

Table \ref{tab:loss_exp} shows the validation accuracy for the various loss functions. We observe at high noise ratio of 0.8 that our proposed projective loss functions performs better by a large margin, with the log-projective loss function performing the best for all noisy settings. For the no noise (i.e. 0 noise ratio), cross-entropy loss performs better, which is evaluated in the following section \ref{noisy_label_exp}.

\begin{table*}[h]
    \centering
    \caption{Validation accuracy on CIFAR10 with Asymmetric noise for various loss functions. $M$ denotes the number of trusted (probe) data used. $1k$ indicates 100 high quality image labels per class. 
    }
    \label{tab:loss_exp}
    \begin{tabular}{llllll}
        \toprule
        Loss Function & $M$ & \multicolumn{4}{c}{Noise ratio}\\
        \cmidrule{3-6} \\[-\normalbaselineskip]
        {} & & 0 & 0.2 & 0.4 & 0.8 \\
        \midrule
        
        CE-RN34 & - & 93.99 $\pm$ 0.47  & 90.13 $\pm$ 0.15 & 85.81 $\pm$ 0.94  & 15.18 $\pm$ 1.48 \\
        pCE-RN34 & 1k & 92.04 $\pm$ 0.42 & 88.50 $\pm$ 0.47 & 72.67 $\pm$ 0.94 & 25.35 $\pm$ 3.45 \\
        L1-RN34 & 1k & 90.63 $\pm$ 0.46 & 89.80 $\pm$ 0.44 & 75.51 $\pm$ 5.34 & 25.13 $\pm$ 1.74 \\
        MSE-RN34 & 1k & 88.57 $\pm$ 1.93 & 84.74 $\pm$ 2.96 & 75.25 $\pm$ 7.42 & 21.02 $\pm$ 0.10 \\
        P-RN34 & 1k & 77.66 $\pm$ 0.88 & 86.58 $\pm$ 5.48 & 78.81 $\pm$ 1.09 & 69.33 $\pm$ 7.47 \\
        LP-RN34 & 1k & 90.53 $\pm$ 0.18 & 90.60 $\pm$ 0.22 & 89.24 $\pm$ 0.09  & 78.13 $\pm$ 0.73 \\
        
        \bottomrule
    \end{tabular}
\end{table*}

\subsubsection{Noisy label experiments}
\label{noisy_label_exp}

We evaluate our log-projection loss for the various noise ratios using trusted set $M$ of 10K, 1K and 0.1K, and we compare its performance with cross-entropy loss.

\subsubsection{Asymmetric noise}



Asymmetric-noise results are presented in Table \ref{tab:asymmetric_exp}. We observe that the log-projection loss does significantly better compared to the baseline cross-entropy loss when using trusted set $M$ of 10k and 1k. With the trusted set of 0.1K the performance drops below that of cross-entropy loss except for 0.8 noise ratio where it is better than cross-entropy. This suggests that the log-projection loss needs a number of high confidence labels to be robust to noise. 

Not surprisingly, for a noise ratio of zero cross-entropy loss performs the best. We observe that the performance of the projection loss at zero noise ratio is far worse compared to the 0.2 noise ratio. It may be that exclusive use of high confidence labels defeats the similarity-preserving capabilities of our approach, resulting in typical spreading of the naturally occurring distribution. So, as the ratio of higher confidence labels is initially reduced from 100\%, our accuracy actually improves over the zero noise ratio.  

The training and test accuracy curves for the log-projection and cross-entropy trained models are shown in figure \ref{fig:training_curves}, at 0.8 noise ratio with trusted set $M$ = 1k. We see that the log-projection model initially starts to overfit on the noisy training set but gradually learns from the higher-confidence labels which features boost accuracy, laying a foundation on which it can learn additional relevant features from the less-confident labels and achieve higher validation accuracy, ultimately ``under fitting'' on the noisy parts of the training set. On the other hand, cross-entropy loss consistently overfits on the noisy train set. \cite{sgd_always_overfit_on_train} found in their experiments that CNN models always overfit on the noisy labels, and \textbf{our log-projection loss is a potential solution to this pervasive problem of overfitting to noisy labels}. 

Figure \ref{fig:class_distribution} shows t-SNE plots \cite{tSNE} of the trained models on the validation set, as done in \cite{tSNE}.  We observe that the cross-entropy-loss trained model with 80\% noise, which overfit on the noisy training set, actually learnt feature embeddings that cluster similar class groups together.  On the other hand, when operating with 0\% noise the cross-entropy model swayed from this natural distribution, though we can still see similar classes are relatively closer. This suggests that all models may have a natural tendency to learn features that cluster according to class similarity. Our proposed Log-Projection loss uniquely embraces this natural distribution when training models, irrespective of the noise setting.

\begin{table*}[h]
    \centering
    \caption{Validation accuracy on CIFAR10 with Asymmetric noise. $M$ denotes the number of trusted (probe) data used. $0.1k$ indicates 10 image per class. 
    }
    \label{tab:asymmetric_exp}
    \begin{tabular}{llllll}
        \toprule
        \multirow{2}{*}{Method} & \multirow{2}{*}{$M$} & \multicolumn{4}{c}{Noise ratio}\\
        \cmidrule{3-6} \\[-\normalbaselineskip]
        {} & & 0 & 0.2 & 0.4  & 0.8 \\
        \midrule
        
        
        DES \cite{google_label_weighting_relabelling} & 0.1k & 96.8 & 96.5 $\pm$ 0.2  & 94.9 $\pm$ 0.1 & 79.3 $\pm$ 2.4 \\
        
        \midrule
        
        CE-RN34 & - & 94.56 $\pm$ 0.16  & 90.13 $\pm$ 0.15 & 85.81 $\pm$ 0.94  & 15.18 $\pm$ 1.48 \\
        LP-RN34 & 10k & 90.26 $\pm$ 0.38 & 92.37 $\pm$ 0.48 & 91.97 $\pm$ 0.50 & 88.51 $\pm$ 1.02 \\
        LP-RN34 & 1k & 90.53 $\pm$ 0.18 & 90.60 $\pm$ 0.22 & 89.24 $\pm$ 0.09  & 78.13 $\pm$ 0.73 \\
        LP-RN34 & 0.1k & 90.49 $\pm$ 0.62 & 88.25 $\pm$ 0.46 & 81.92 $\pm$ 2.14 & 30.18 $\pm$ 5.27 \\
        
        \bottomrule
    \end{tabular}
\end{table*}

\subsubsection{Symmetric noise}


Symmetric-noise results are presented in Table \ref{tab:symmetric_exp}. Our log-projection loss was not designed for symmetric noise, and we observe that it does not do as well as it did in the asymmetric noise case. Log-projection loss does achieve comparable (often better) performance to baseline cross-entropy loss when using a large ratio of high confidence labels ($M$ = 10k). It may be that our projection loss struggles to exploit class similarity relations in uninformative symmetric noise, requiring a large number of high confidence labels to discover and utilize the underlying similarities.  Google's label-weighting/relabeling approach \cite{google_label_weighting_relabelling}, however, was the best performer for symmetric label noise. 

\begin{table*}[h]
    \centering
    \caption{Validation accuracy on CIFAR10 with Symmetric noise. $M$ denotes the number of trusted (probe) data used. $0.1k$ indicates 10 images per class. 
    }
    \label{tab:symmetric_exp}
    \begin{tabular}{llllll}
        \toprule
        \multirow{2}{*}{Method} & \multirow{2}{*}{$M$} & \multicolumn{4}{c}{Noise ratio}\\
        \cmidrule{3-6} \\[-\normalbaselineskip]
        {} & & 0 & 0.2 & 0.4  & 0.8 \\
        \midrule
        
        
        DES \cite{google_label_weighting_relabelling} & 0.1k & 96.8 & 96.3 $\pm$ 0.2  & 95.9 $\pm$ 0.3 & 93.7 $\pm$ 0.5 \\
        
        \midrule
        
        
        CE-RN34 & - & 94.56 $\pm$ 0.16  & 88.75 $\pm$ 0.19 & 83.84 $\pm$ 1.03  & 41.52 $\pm$ 3.31 \\
        LP-RN34 & 10k & 90.26 $\pm$ 0.38 & 88.78 $\pm$ 0.28 & 84.15 $\pm$ 1.17 & 48.05 $\pm$ 5.06 \\
        LP-RN34 & 1k & 90.53 $\pm$ 0.18 & 88.52 $\pm$ 0.22 &  84.11 $\pm$ 0.35 &  32.03 $\pm$ 5.60 \\
        LP-RN34 & 0.1k & 90.49 $\pm$ 0.62 & 87.82 $\pm$ 1.04 & 83.32 $\pm$ 0.33 & 29.79 $\pm$ 3.83 \\
        
        \bottomrule
    \end{tabular}
\end{table*}

\subsection{Large-scale real-world experiments}
\label{real-world_exp}

In this section, we experiment on Food101N and ImageNet datasets, which are both large-scale datasets. We show two approaches of our method: (1) a way to automatically generate confidence labels using pretrained models in order to refine existing model performance, in section \ref{imagenet} and (2) we show applicability of our approach in a straightforward setup without the need of generating confidence labels by directly operating on one-hot labels, in section \ref{food101n}. 

\subsubsection{Implementation details and experimental setup}

For both the experiments on ImageNet and Food101N, we evaluate using ResNet-50 implemented with PyTorch and trained using the stochastic gradient descent algorithm \cite{stochasticGradientUpdate} with a weight decay of $5e-4$, set with an initial learning rate of $0.005$ and $0.0005$ for Food101N and ImageNet experiments respectively. The model is trained on 10 Nvidia GeForce RTX 2080 Ti GPUs with a batch size of 1000 for gradient update computation. The ReduceLRonPlateau learning-rate scheduler was used which reduces the learning rate by a factor (0.5) when the performance accuracy plateaus on the test set, with patience 1. The model is trained for 50 and 20 epochs for Food101N and ImageNet experiments respectively. For the final evaluation, we pick the best model with the highest test set accuracy. Similar to experiments in section \ref{cifar_exps} we only make use of normal augmentaion with image resized to 224x224 and do not make use of other standard training-enhancement techniques. 



\subsubsection{Fine-tuning existing models training}
\label{imagenet}



An automated way of generating confidence labels is to use the outputs of the existing trained neural network architectures by making use of the trained model's prediction (softmax last layer output). A weighted average of multiple trained models can be taken to generate confidence labels wherein the weights are based on the models performance accuracy. This method would be useful when dealing with large number of target classes such as ImageNet dataset which has 1000 label classes, wherein manual confidence labelling would be cumbersome/unwieldy. 

We demonstrate this on ImageNet by using the model's prediction of a trained ResNet-50 model \footnote{Trained ResNet-50 model weights obtained from https://pytorch.org/}. We define the similarity score $S_a$ for the class $a$ as the average of all the model predictions with class label $a$. For each class $a$, we choose its threshold $\tau_a$ to be the mean plus one standard deviation of the similarity scores $S_a$.  We thus obtain confidence labels $C_a$ for all classes as defined in \ref{eq:C_a}. 

We further train the model initialized with the pretrained weights using these generated confidence labels using the log-projection loss. We use the non-noisy relaxation function defined in \ref{eq:no_relax} ($r(t) = t$, i.e. no relaxation), as ImageNet is predominantly a non-noisy dataset. We do not make use of any trusted set $M$ for these experiments. 

The results are shown in Table \ref{tab:imagenet}.  We observe that the baseline trained model's accuracy is further improved by training the model using confidence labels and log-projection loss. We nevertheless suspect that the generated confidence labels were not representative of the similarity distribution, as the model used to generate the labels had only 76.15\%  accuracy. Since ImageNet is a non-noisy dataset, we carry out another log-projection training experiment using the the one-hot labels as the confidence labels, as  shown in the table.
We see substantial improvements with this, comparable to improvements obtained using AutoAugement \cite{autoaugment}. The observed improvements are without incorporating many ``standard'' training-enhancement techniques, so further gains are presumably possible.







\begin{table}[h]
    \centering
    \caption{Validation accuracy on ImageNet. 
    }
    \label{tab:imagenet}
    \begin{tabular}{llll}
        \toprule
        Method & Label Type & Accuracy \\
        
        \midrule
        AutoAugment \cite{autoaugment} & CE-RN50 & one-hot & 77.6 \\
         
        \midrule
        baseline & RN50 & one-hot & 76.15 \\
        Ours & LP-RN50 & confidence & 76.59 \\
        Ours & LP-RN50 & one-hot & 76.92 \\
        \bottomrule
    \end{tabular}
\end{table}

\subsubsection{Food101N dataset}
\label{food101n}

In this experiment, we directly apply our projective loss function on one-hot labels, instead of on confidence labels.
We follow the experimental setup of \cite{food101n_setup_han2019deep}, wherein we train the model using ImageNet pretrained weights, on all the images of Food-101N dataset and evaluate on the test set of Food-101 dataset \cite{food101_dataset}. We make use of the normal one-hot labels as the confidence labels themselves. The L2-norm heuristic function $H$ applied on the one-hot labels will have a value of 1, signifying higher confidence on the labels. We do not use any trusted set $M$ for these experiments.

The experimental results are reported in Table \ref{tab:food101n}. We observe that our approach achieves comparable performance to that of the current state-of-the-art technique, even when directly using the one-hot labels as the confidence labels, and without any loss relaxation. We note that directly training the model without pretrained ImageNet weights achieved lower accuracy. We therefore show that using our projective loss function does not hurt model performance on large-scale real-world datasets, which are inherently noisy. 


\begin{table}[h]
    \centering
    \caption{Food101N experiments.
    }
    \label{tab:food101n}
    \begin{tabular}{lll}
        \toprule
        Method & Accuracy \\
        
        \midrule
        CleanNet \cite{food101n_dataset_lee2017cleannet} & & 83.59 \\
        Self-Learning \cite{food101n_setup_han2019deep} & & 85.11 \\
        DES \cite{google_label_weighting_relabelling} & CE-RN50 & 87.57 \\
        
        \midrule
        
        Ours & LP-RN50 & 84.97 \\
        \bottomrule
    \end{tabular}
\end{table}

\section{Conclusion}


We presented confidence labels, which encode a-priori inter-class similarity relations in order to supplement model training with class similarity relations. We also introduced a family of projective loss functions that preserve the naturally occurring class similarity distributions, tailored to work with confidence labels and with the ability to relax the loss penalty. We showed that our simple approach achieves comparable performance to the state-of-the-art methods in the asymmetric noise setting. We also showed potential improvements to be gained using our approach on real-world large-scale ImageNet and Food-101N datasets, without yet incorporating many of the ``standard'' training-enhancement techniques. We plan to extend our method to open-set learning and few-shot learning scenarios.





{\small
\bibliographystyle{ieee_fullname}
\bibliography{main}
}

\end{document}